\documentclass[10pt,twocolumn,letterpaper]{article}

\usepackage{sty/cvpr}
\usepackage{times}
\usepackage{epsfig}
\usepackage{graphicx}
\usepackage{amsmath}
\usepackage{amssymb}

\usepackage{url}  %
\usepackage{graphicx, color, xspace}
\usepackage{subcaption}
\usepackage{booktabs}
\usepackage{multirow}

\usepackage[breaklinks=true,colorlinks,bookmarks=false]{hyperref}

\cvprfinalcopy %

\def\cvprPaperID{2124} %

\ifcvprfinal\fi
\begin{document}

\title{MnasNet: Platform-Aware Neural Architecture Search for Mobile}

\author{
	\small
	\begin{tabular}{c c c c c c c }                                                          
		\bf Mingxing Tan$^1$ &
		\bf Bo Chen$^2$ &
		\bf Ruoming Pang$^1$ &
		\bf Vijay Vasudevan$^1$ &
		\bf Mark Sandler$^2$ &
		\bf Andrew Howard$^2$ &
		\bf Quoc V. Le$^1$ \\                                        
		\multicolumn{7}{c}{$^1$Google Brain, $^2$Google Inc.} \\                                                  
		\multicolumn{7}{c}{\{\tt\small tanmingxing, bochen, rpang, vrv, sandler, howarda, qvl\}@google.com} \\
	\end{tabular}                                                                       
}   

\maketitle
\thispagestyle{empty}

\def\TODO{\textcolor{red}{\emph{TODO: }}}
\newcommand{\TT}[1]{\emph{#1}}
\newcommand{\BF}[1]{\textbf{#1}}
\newcommand{\IT}[1]{\textit{#1}}

\newcommand\blfootnote[1]{%
	\begingroup
	\renewcommand\thefootnote{}\footnote{#1}%
	\addtocounter{footnote}{-1}%
	\endgroup
} 
\begin{abstract}
Designing convolutional neural networks (CNN)  for mobile devices is challenging because mobile models need to be small and fast, yet still accurate. Although significant efforts have been dedicated to design and improve mobile CNNs on all dimensions, it is very difficult to manually balance these trade-offs when there are so many architectural possibilities to consider. In this paper, we propose an  automated  mobile neural architecture search (MNAS) approach, which explicitly incorporate model latency  into the main objective so that the search can identify a model that achieves a good trade-off between accuracy and latency. Unlike previous work, where latency is considered via another, often inaccurate proxy (e.g., FLOPS), our approach directly measures real-world inference latency by executing the model on mobile phones.
To further strike the right balance between flexibility and search space size, we propose a novel factorized hierarchical search space that encourages layer diversity throughout the network. Experimental results show that our approach consistently outperforms state-of-the-art mobile CNN models across multiple vision tasks. On the ImageNet classification task, our MnasNet achieves 75.2\% top-1 accuracy with 78ms latency on a Pixel phone, which is $1.8\times$ faster than MobileNetV2 \cite{mobilenetv218} with 0.5\% higher accuracy and $2.3\times$ faster than NASNet \cite{nas_imagenet18} with 1.2\% higher accuracy. Our MnasNet also achieves better mAP quality than MobileNets for COCO object detection. Code is at \url{https://github.com/tensorflow/tpu/tree/master/models/official/mnasnet}.
\end{abstract}
\section{Introduction}
\label{sec:intro}

\begin{figure}
	\centering
	\includegraphics[width=.9\columnwidth]{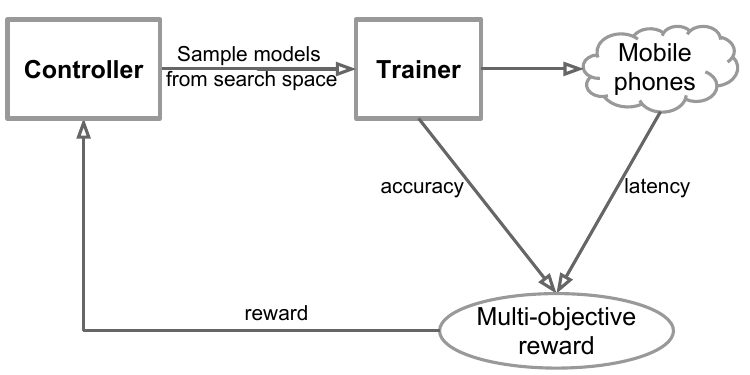}
	\caption{\textbf{An Overview of Platform-Aware Neural Architecture Search for Mobile}. 
	}
	\label{fig:overall}
\end{figure} %
\begin{figure}
	\centering
	\includegraphics[width=.95\columnwidth]{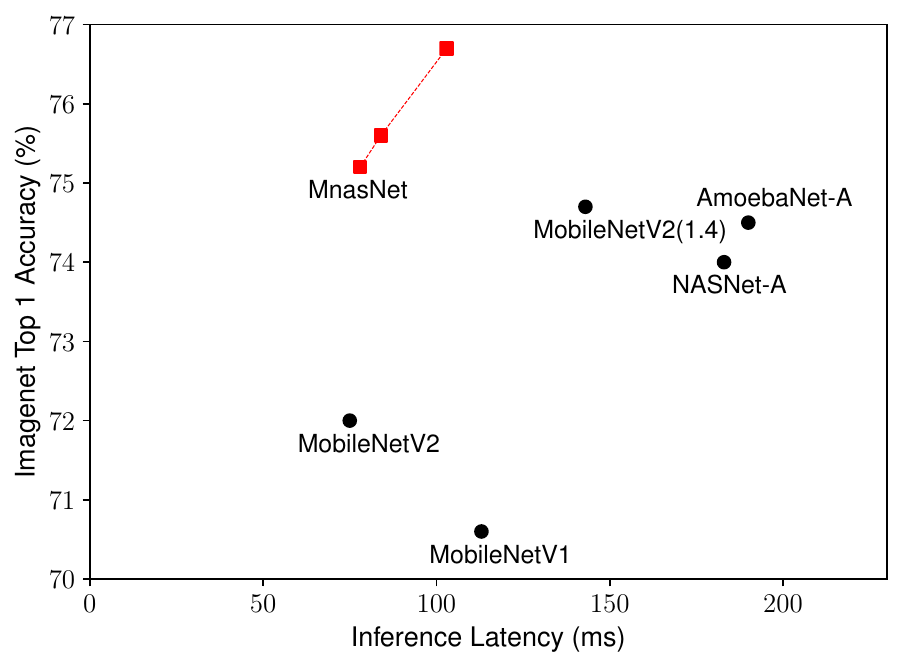}
	\caption{\textbf{Accuracy vs. Latency Comparison} -- Our MnasNet models significantly outperforms other mobile models \cite{mobilenetv218,nas_imagenet18,amoebanets18} on ImageNet. Details can be found in Table \ref{tab:imagenet}.
	}
	\label{fig:mobilelatency}
\end{figure} 
Convolutional neural networks (CNN) have made significant progress in image classification, object detection, and many other applications. As modern CNN models  become increasingly deeper and larger \cite{inceptionresnet17,senet18,nas_imagenet18,amoebanets18}, they also become slower, and require more computation. Such increases in computational demands  make it difficult to deploy state-of-the-art CNN models on resource-constrained platforms such as mobile or embedded devices.

Given restricted computational resources available on mobile devices, much recent research has focused on designing and improving mobile CNN models by reducing the depth of the network and utilizing less expensive operations, such as depthwise convolution \cite{mobilenetv117} and group convolution \cite{shufflenet17}. However, designing a resource-constrained mobile model is challenging: one has to carefully balance accuracy and resource-efficiency, resulting in a significantly large design space.

In this paper, we propose an automated neural architecture search approach for designing mobile CNN models. Figure \ref{fig:overall} shows an overview of our approach, where the main differences from previous approaches are the latency aware multi-objective reward and the novel search space.
Our approach is based on two main ideas. 
First, we formulate the design problem as a multi-objective optimization problem that considers both accuracy and inference latency of CNN models. Unlike in previous work \cite{nas_imagenet18,amoebanets18,diffnas18} that use FLOPS to approximate inference latency, we directly measure the real-world  latency by executing the model on real mobile devices. Our idea is inspired by the observation 
that FLOPS is often an inaccurate proxy: for example, MobileNet \cite{mobilenetv117} and NASNet \cite{nas_imagenet18} have similar FLOPS (575M vs. 564M), but their latencies are significantly different (113ms vs. 183ms, details in Table \ref{tab:imagenet}).
Secondly, we observe that previous automated approaches mainly search for a few types of cells and then repeatedly stack the same cells through the network. This simplifies the search process, but also precludes layer diversity that is important for computational efficiency. To address this issue, we propose a novel \emph{factorized hierarchical search space}, which allows layers to be architecturally different yet still strikes the right balance between flexibility and search space size.

We apply our proposed approach to ImageNet classification \cite{imagenet15} and COCO object detection \cite{coco14}.
Figure \ref{fig:mobilelatency} summarizes a comparison between our MnasNet models and other state-of-the-art mobile models.
Compared to the MobileNetV2 \cite{mobilenetv218}, our model improves the ImageNet accuracy by 3.0\% with similar latency on the Google Pixel phone. On the other hand, if we constrain the target accuracy, then our MnasNet models are  \textbf{$\mathbf{1.8\times}$ faster} than MobileNetV2 and \textbf{$\mathbf{2.3\times}$ faster} thans NASNet \cite{nas_imagenet18} with better accuracy. 
Compared to the widely used ResNet-50 \cite{resnet16}, our MnasNet model achieves slightly higher (76.7\%)  accuracy with \textbf{$\mathbf{4.8\times}$ fewer} parameters and \textbf{$\mathbf{10\times}$ fewer} multiply-add operations.
By plugging our model as a  feature extractor into the SSD object detection framework, our model improves both the inference latency and the mAP quality on COCO dataset over MobileNetsV1 and MobileNetV2, and achieves comparable mAP quality (23.0 vs 23.2) as SSD300 \cite{ssd16} with \textbf{$\mathbf{42\times}$ less}  multiply-add operations.

To summarize, our main contributions are as follows:

\begin{enumerate}
	\item We introduce a \emph{multi-objective}  neural architecture search approach that optimizes both accuracy and real-world latency on mobile devices.
	
	\item We propose a novel \emph{factorized hierarchical search space} to enable layer diversity yet still strike the right balance between  flexibility and search space size.
	
	\item We demonstrate new state-of-the-art accuracy on both ImageNet classification and COCO object detection under typical mobile latency constraints.
\end{enumerate} %
\section{Related Work}
\label{sec:related}

Improving the resource efficiency of CNN models has been an active research topic during the last several years. Some commonly-used approaches include 1) quantizing the weights and/or activations of a baseline CNN model into lower-bit representations \cite{quantize15,fackequantize18}, or 2)  pruning less important filters according to FLOPs  \cite{morphnet18,amc18}, or to platform-aware metrics such as latency introduced in \cite{netadapt18}. However, these methods are tied to a baseline model and do not focus on learning novel compositions of CNN operations. 

Another common approach is to directly hand-craft more efficient mobile architectures: SqueezeNet \cite{squeezenet16} reduces the number of parameters and computation by using lower-cost 1x1 convolutions and reducing filter sizes; MobileNet \cite{mobilenetv117} extensively employs depthwise separable convolution to minimize computation density; ShuffleNets \cite{shufflenet17, shufflenetv218} utilize low-cost group convolution and channel shuffle; Condensenet \cite{condensenet18} learns to connect group convolutions across layers; Recently, MobileNetV2 \cite{mobilenetv218} achieved state-of-the-art results among mobile-size models by using resource-efficient inverted residuals and linear bottlenecks. Unfortunately, given the potentially huge design space, these hand-crafted models usually take significant human efforts.

Recently, there has been growing interest in automating the model design process  using neural architecture search. These approaches are mainly based on reinforcement learning \cite{nas_cifar17,nas_imagenet18,metaqnn17,pnas18,enas18}, evolutionary search \cite{amoebanets18}, differentiable search  \cite{diffnas18}, or other learning algorithms \cite{pnas18,bayesian2018neural,nao18neural}. Although these methods can generate mobile-size models by repeatedly stacking a few searched cells, they do not incorporate mobile platform constraints into the search process or search space. Closely related to our work is MONAS \cite{monas18}, DPP-Net \cite{dppnet18}, RNAS \cite{rnas18} and Pareto-NASH \cite{multiobj18} which attempt to optimize multiple objectives, such as model size and accuracy, while searching for CNNs, but their search process optimizes on small tasks like CIFAR.  In contrast, this paper targets real-world mobile latency constraints and focuses on larger tasks like ImageNet classification and COCO object detection.  %
\section{Problem Formulation}
\label{sec:problem}

\begin{figure}
	\centering
	\includegraphics[width=.98\columnwidth]{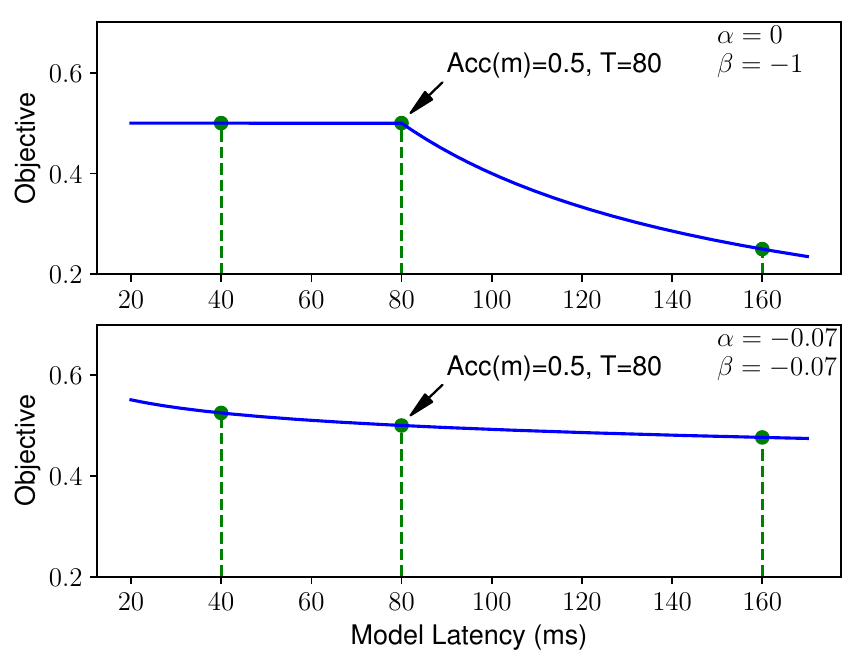}
	\caption{\textbf{Objective Function Defined by Equation \ref{eq:reward}}, assuming accuracy $ACC(m)$=0.5 and target latency $T$=80ms: (top) show the object values with latency as a hard constraint; (bottom) shows the objective values with latency as a soft constraint.
	}
	\label{fig:reward}
\end{figure} 
We formulate the design problem as a multi-objective search, aiming at finding CNN models with both high-accuracy and low inference latency. Unlike  previous architecture search approaches that often optimize for indirect metrics, such as FLOPS, we consider direct \emph{real-world inference latency},  by running CNN models on real mobile devices,  and then incorporating the real-world inference latency into our objective. Doing so directly measures what is achievable in practice:  our early experiments show it is challenging to approximate real-world latency due to the variety of mobile hardware/software idiosyncrasies.

Given a model $m$, let $ACC(m)$  denote its accuracy on the target task, $LAT(m)$ denotes the inference latency on the target mobile platform, and $T$ is the target latency.  A common method is to treat $T$ as a hard constraint and maximize accuracy under this constraint:

\begin{equation} \label{eq:opthard}
\begin{aligned}
& \underset{m}{\text{maximize }} & & ACC(m) \\
& \text{subject to} & & LAT(m) \le T
\end{aligned}
\end{equation} 

\noindent However, this approach only maximizes a single metric and does not provide multiple Pareto optimal solutions. Informally, a model is called Pareto optimal \cite{multiobjbook14} if either it has the highest accuracy without increasing latency or it has the lowest latency without decreasing accuracy. Given the computational cost of performing architecture search, we are more interested in finding multiple Pareto-optimal solutions in a single architecture search.

While there are many methods in the literature \cite{multiobjbook14}, we use a customized weighted product method\footnote{We pick the weighted product method because it is easy to customize, but we expect methods like weighted sum should be also fine.} to approximate Pareto optimal solutions, with  optimization goal defined as:

\begin{equation} \label{eq:reward}
	\begin{aligned}
	& \underset{m}{\text{maximize  }} & & {ACC(m)} \times \left[ \frac{ LAT(m)}{T} \right] ^ w
	\end{aligned}
\end{equation}

\noindent where $w$ is the weight factor defined as:

\begin{equation} \label{eq:alpha}
\begin{aligned}
w = 
\begin{cases}
\alpha,               & \text{if } LAT(m) \leq T\\
\beta,                & \text{otherwise}
\end{cases}
\end{aligned}
\end{equation} 

\noindent where $\alpha$ and $\beta$ are application-specific constants. An empirical rule for picking $\alpha$ and $\beta$ is to
ensure Pareto-optimal solutions have similar reward under different accuracy-latency trade-offs. For instance, we empirically observed doubling the latency usually brings about 5\% relative accuracy gain. Given two models: (1) M1 has latency $l$ and accuracy $a$; (2) M2 has latency $2l$ and 5\% higher accuracy $ a \cdot (1 + 5\%)$, they should have similar reward: $Reward(M2) = a \cdot (1+ 5\%) \cdot (2l / T)^\beta \approx Reward(M1) = a \cdot (l / T)^\beta$. Solving this gives $\beta \approx -0.07$. Therefore, we use $\alpha=\beta=-0.07$ in our experiments unless explicitly stated.

Figure \ref{fig:reward} shows the objective function with two typical values of $(\alpha, \beta)$. In the top figure with ($\alpha=0, \beta=-1$), we simply use accuracy as the objective value if measured latency is less than the target latency $T$; otherwise, we sharply penalize the objective value to discourage models from violating latency constraints.  The bottom figure ($\alpha=\beta=-0.07$) treats the target latency $T$ as a soft constraint, and smoothly adjusts the objective value based on the measured latency. 
\section{Mobile Neural Architecture Search}
\label{sec:mnas}

In this section, we will first discuss our proposed novel factorized hierarchical search space, and then summarize our reinforcement-learning based search algorithm.

\begin{figure*}[!htb]
	\centering
	\includegraphics[width=0.93\textwidth]{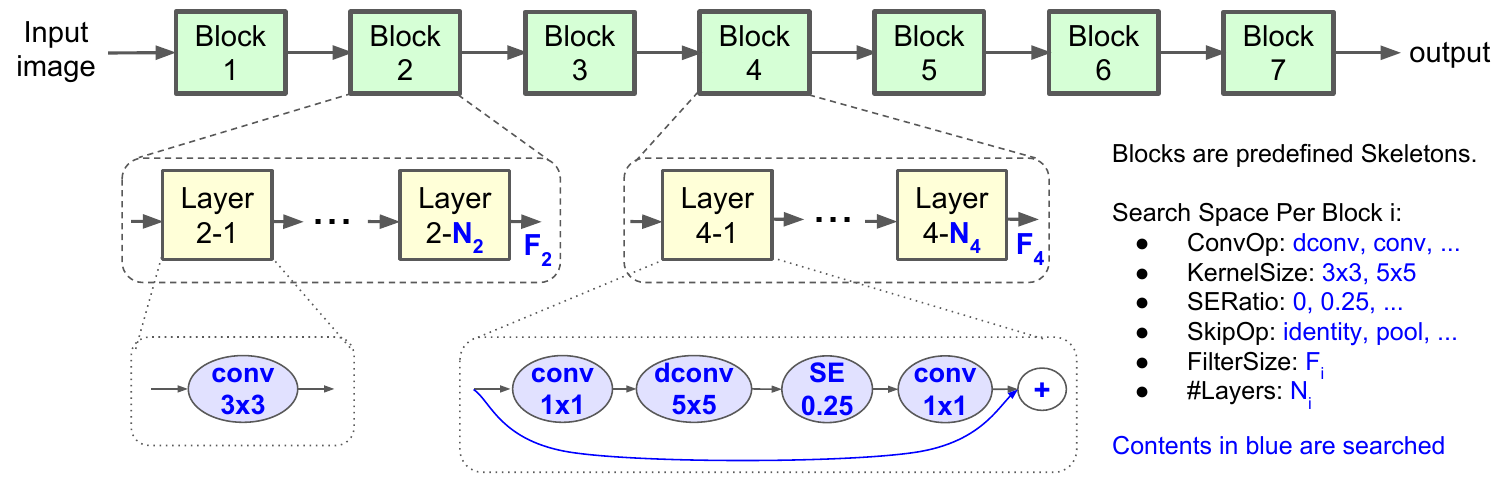}
	\caption{\textbf{Factorized Hierarchical Search Space.}
		Network layers are grouped into a number of predefined skeletons, called blocks, based on their input resolutions and filter sizes. Each block contains a variable number of repeated identical layers 
		where only the first layer has stride 2 if input/output resolutions are different but all other layers have stride 1. For each block, we search for the operations and connections for a single layer and the number of layers $N$, then the same layer is repeated $N$ times (e.g., Layer 4-1 to 4-N$_4$ are the same). Layers from different blocks (e.g., Layer 2-1 and 4-1) can be different.
	}
	\label{fig:searchspace}
\end{figure*} 
\subsection{Factorized Hierarchical Search Space}
\label{subsec:searchspace}
As shown in recent studies \cite{nas_imagenet18,hnas18}, a well-defined search space is extremely important for neural architecture search. However, most previous  approaches \cite{nas_cifar17,pnas18,amoebanets18} only search for a few complex cells and then repeatedly stack the same cells. These approaches don't permit layer diversity, which we show is critical for achieving both high accuracy and lower latency.

In contrast to previous approaches, we introduce a novel {factorized hierarchical search space} that factorizes a CNN model into unique blocks and then searches for the operations and connections per block separately, thus allowing different layer architectures in different blocks.
Our intuition is that we need to search for the best operations based on the input and output shapes to obtain better accurate-latency trade-offs. For example, earlier stages of CNNs usually process larger amounts of data and thus have much higher impact on inference latency than later stages. Formally, consider a widely-used depthwise separable convolution \cite{mobilenetv117} kernel denoted as the four-tuple $(K, K,  M, N)$ that transforms an input of size $(H, W, M)$\footnote{We omit batch size dimension for simplicity.} to an output of size $(H, W, N)$, where $(H, W)$ is the input resolution and $M, N$ are the input/output filter sizes. The total number of multiply-adds  can be described as:

\begin{equation}
H * W * M * (K * K  +  N)
\end{equation}

\noindent
Here we need to carefully balance the kernel size $K$ and filter size $N$ if the total computation is constrained. For instance, increasing the receptive field with larger kernel size $K$ of a layer must be balanced with reducing either the filter size $N$ at the same layer, or compute from other layers.

Figure \ref{fig:searchspace} shows the baseline structure of our search space. We partition a CNN model into a sequence of pre-defined blocks, gradually reducing  input resolutions and increasing filter sizes as is common in many CNN models. Each block has a list of identical layers, whose operations and connections are determined by a per-block sub search space. Specifically, a sub search space for a block $i$ consists of the following choices:

\begin{itemize}
	\small	
	 \setlength\itemsep{0em}
	\item Convolutional ops $ConvOp$: regular conv (conv), depthwise conv (dconv), and mobile inverted bottleneck conv \cite{mobilenetv218}.
	\item Convolutional kernel size $KernelSize$: 3x3, 5x5.
	\item Squeeze-and-excitation \cite{senet18} ratio $SERatio$: 0, 0.25.
	\item Skip ops $SkipOp$:  pooling, identity residual, or no skip.
	\item Output filter size $F_i$.
	\item Number of layers per block $N_i$.
\end{itemize}

\noindent
$ConvOp$, $KernelSize$, $SERatio$, $SkipOp$, $F_i$ determines the architecture of a layer, while $N_i$ determines how many times the layer will be repeated for the block. For example, each layer of block 4 in Figure \ref{fig:searchspace} has an inverted bottleneck 5x5 convolution and an identity residual skip path, and the same layer is repeated $N_4$ times. We discretize all search options using MobileNetV2 as a reference: For \#layers in each block, we search for \{0, +1, -1\} based on MobileNetV2; for filter size per layer, we search for its relative size in \{0.75, 1.0, 1.25\} to MobileNetV2 \cite{mobilenetv218}.

Our factorized hierarchical search space has a distinct advantage of balancing the diversity of layers and the size of total search space. Suppose we partition the network into $B$ blocks, and each block has a sub search space of size $S$ with average $N$ layers per block, then our total search space size would be $S^B$, versing the flat per-layer search space with size $S ^{B*N}$. A typical case is $S=432, B=5, N=3$, where our search space size is about $10^{13}$, versing the per-layer approach with search space size $10^{39}$.

\begin{table*}
    \centering 
    \resizebox{0.98\textwidth}{!}{
        \begin{tabular}{l|c|cc|cc|c}
        \toprule [0.2em]
        Model & Type &  \#Params & \#Mult-Adds & Top-1 Acc. (\%) & Top-5 Acc. (\%) & Inference Latency \\
        \toprule [0.2em]
        MobileNetV1  \cite{mobilenetv117}     & manual  & 4.2M  & 575M & 70.6 & 89.5 & 113ms  \\ 
        SqueezeNext  \cite{squeezeNext18}    & manual  & 3.2M  & 708M  & 67.5 & 88.2 & - \\
        ShuffleNet (1.5x)  \cite{shufflenet17} & manual  & 3.4M  & 292M & 71.5 & - & - \\
        ShuffleNet (2x)                       & manual  & 5.4M  & 524M & 73.7 & - & - \\
        ShuffleNetV2 (1.5x) \cite{shufflenetv218}   & manual  & -  & 299M & 72.6 & - & - \\
        ShuffleNetV2 (2x)   & manual  & -  & 597M & 75.4 & - & - \\
        CondenseNet (G=C=4) \cite{condensenet18}    & manual  &  2.9M  & 274M & 71.0 & 90.0 & - \\
        CondenseNet (G=C=8)    & manual  &  4.8M  & 529M & 73.8 & 91.7 & - \\
        MobileNetV2  \cite{mobilenetv218}     & manual  & 3.4M  & 300M & 72.0 & 91.0 & 75ms \\ 
        MobileNetV2 (1.4x)    & manual  & 6.9M  & 585M & 74.7 & 92.5 & 143ms \\
        \toprule[0.02em]
        NASNet-A      \cite{nas_imagenet18}   & auto    & 5.3M  & 564M & 74.0 & 91.3 & 183ms  \\
        AmoebaNet-A   \cite{amoebanets18}     & auto    & 5.1M  & 555M & 74.5 & 92.0 & 190ms \\
        PNASNet      \cite{pnas18}    & auto    & 5.1M  & 588M & 74.2 & 91.9 & - \\
        DARTS           \cite{diffnas18}         & auto    & 4.9M  & 595M & 73.1 & 91    & -  \\
        \toprule[0.2em]
        \bf MnasNet-A1                                &  \bf auto    & \bf  3.9M  & \bf 312M & \bf 75.2 & \bf 92.5 & \bf 78ms \\
        \bf MnasNet-A2                                &  \bf auto    & \bf 4.8M  & \bf 340M & \bf 75.6 & \bf 92.7 & \bf 84ms \\
        \bf MnasNet-A3                               &  \bf auto    &  \bf 5.2M  & \bf 403M & \bf 76.7 & \bf 93.3 & \bf 103ms  \\
        \toprule[0.2em]
        \end{tabular}
    }
    \caption{
        \textbf{Performance Results on ImageNet Classification} \cite{imagenet15}. We compare our MnasNet models with both manually-designed mobile models and other automated approaches -- \TT{MnasNet-A1} is our baseline model;\TT{MnasNet-A2} and \TT{MnasNet-A3} are two models (for comparison) with different latency from the same architecture search experiment; \TT{\#Params}: number of trainable parameters; \TT{\#Mult-Adds}: number of multiply-add operations per image; \TT{Top-1/5 Acc.}: the top-1 or top-5 accuracy on ImageNet validation set; \TT{Inference Latency} is measured on the big CPU core of a Pixel 1 Phone with batch size 1.
    }
    \label{tab:imagenet}
\end{table*}

\subsection{Search Algorithm}
Inspired by recent work \cite{nas_cifar17,nas_imagenet18,enas18,hnas18}, we use a reinforcement learning approach to find Pareto optimal solutions for our multi-objective search problem. We choose reinforcement learning because it is convenient and the reward is easy to customize, but we expect other methods like evolution \cite{amoebanets18} should also work. 

Concretely, we follow the same idea as \cite{nas_imagenet18} and map each CNN model in the search space to a list of tokens. These tokens are determined by a sequence of actions $a_{1:T}$ from the reinforcement learning agent based on its parameters $\theta$. Our goal is to maximize the expected reward:

\begin{equation} \label{eq:searchobject}
\begin{aligned}
J=E_{P(a_{1:T}; \theta)} [R(m)]
\end{aligned}
\end{equation}

\noindent
where $m$ is a sampled model  determined by action $a_{1:T}$, and $R(m)$ is the objective value defined by equation \ref{eq:reward}.

As shown in Figure \ref{fig:overall}, the search framework consists of three components: a recurrent neural network (RNN) based controller, a trainer to obtain the model accuracy, and a mobile phone based inference engine for measuring the latency. We follow the well known sample-eval-update loop to train the controller. At each step, the controller first samples a batch of models using its current parameters $\theta$, by predicting a sequence of tokens based on the softmax logits from its RNN. For each sampled model $m$, we train it on the target task to get its accuracy $ACC(m)$, and run it on real phones to get its inference latency $LAT(m)$. We then calculate the reward value $R(m)$ using equation \ref{eq:reward}. At the end of each step, the parameters $\theta$ of the controller are updated by maximizing the expected reward defined by equation \ref{eq:searchobject} using Proximal Policy Optimization~\cite{schulman2017proximal}. The sample-eval-update loop is repeated until it reaches the maximum number of steps or the parameters $\theta$ converge.
\begin{figure*}
    \centering
    \begin{subfigure}{0.45\textwidth}                                               
        \centering                                                                  
        \includegraphics[width=\linewidth,keepaspectratio=true]{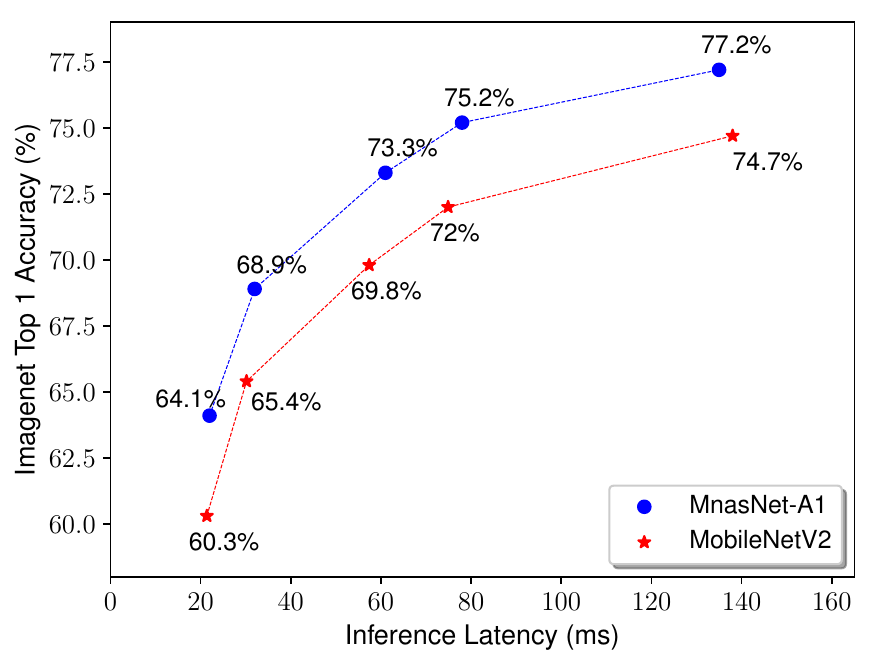}
        \caption{Depth multiplier = 0.35, 0.5, 0.75, 1.0, 1.4, corresponding to points from left to right.}                  
        \label{fig:filter}
    \end{subfigure}
    \hspace{.5in}
    \begin{subfigure}{0.45\textwidth}
		\centering                                                                  
		\includegraphics[width=\linewidth,keepaspectratio=true]{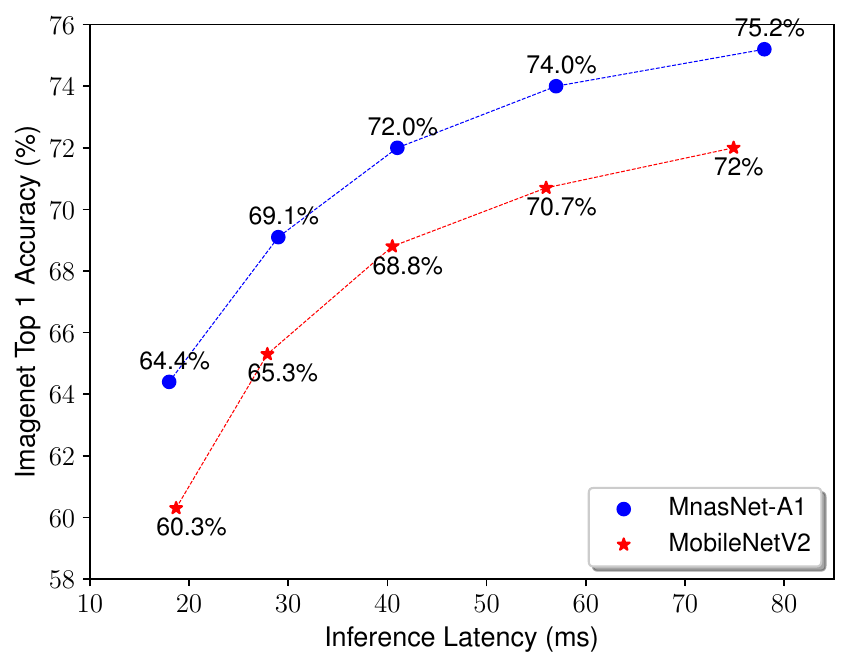}
		\caption{Input size = 96, 128, 160, 192, 224, corresponding to points from left to right.}                  
		\label{fig:input}
   \end{subfigure}
    \caption{
        \textbf{Performance Comparison with Different Model Scaling Techniques}.  MnasNet is our baseline model shown in Table \ref{tab:imagenet}. We scale it with the same depth multipliers and input sizes as MobileNetV2.
    }
    \label{fig:filter-input}
\end{figure*} 
\section{Experimental Setup}
\label{sec:setups}

Directly searching for CNN models on large tasks like ImageNet or COCO is expensive, as each model takes days to converge. While previous approaches mainly perform architecture search on smaller tasks such as CIFAR-10 \cite{nas_imagenet18,amoebanets18}, we find those small proxy tasks don't work when model latency is taken into account, because one typically needs to scale up the model when applying to larger problems. In this paper, we directly perform our architecture search on the ImageNet training set but with fewer training steps (5 epochs). As a common practice, we reserve randomly selected 50K images from the training set as the fixed validation set. 
To ensure the accuracy improvements are from our
search space, we use the same RNN controller as NASNet \cite{nas_imagenet18} even though it is not efficient: each architecture search takes ~4.5 days on 64 TPUv2 devices. During training, we measure the real-world latency of each sampled model by running it on the single-thread big CPU core of Pixel 1 phones. In total, our controller samples about 8K models during architecture search, but only 15 top-performing models are transferred to the full ImageNet and only 1 model is transferred to  COCO. 

For full ImageNet training, we use RMSProp optimizer with decay 0.9 and momentum 0.9. Batch norm is added after every convolution layer with momentum 0.99, and weight decay is 1e-5. Dropout rate 0.2 is applied to the last layer. Following \cite{imagenet1hour17}, learning rate is increased from 0 to 0.256 in the first 5 epochs, and then decayed by 0.97 every 2.4 epochs.
We use batch size 4K and  Inception preprocessing with image size $224 \times 224$. For COCO training, we plug our learned model into SSD detector \cite{ssd16} and use the same settings as \cite{mobilenetv218}, including  input size $320\times 320$.

\section{Results}
\label{sec:results}

In this section, we study the performance of our models on ImageNet classification and COCO object detection, and compare them with other state-of-the-art mobile models.

\subsection{ImageNet Classification Performance}
Table \ref{tab:imagenet} shows the performance of our models on ImageNet \cite{imagenet15}. We set our target latency as $T=75ms$, similar to MobileNetV2 \cite{mobilenetv218}, and use Equation \ref{eq:reward} with $\alpha$=$\beta$=-0.07 as our reward function during architecture search. Afterwards, we pick three top-performing MnasNet models, with different latency-accuracy trade-offs from the same search experiment and compare them with existing mobile  models.

As shown in the table, our MnasNet A1 model achieves 75.2\% top-1 / 92.5\% top-5 accuracy with 78ms latency and 3.9M parameters / 312M multiply-adds, achieving a new state-of-the-art accuracy for this typical mobile latency constraint. In particular, MnasNet runs \textbf{$\mathbf{1.8\times}$ faster} than MobileNetV2 (1.4) \cite{mobilenetv218} on the same Pixel phone with 0.5\% higher accuracy.  Compared with automatically searched CNN models, our MnasNet runs \textbf{$\mathbf{2.3\times}$ faster} than the mobile-size NASNet-A \cite{nas_imagenet18} with 1.2\% higher top-1 accuracy. Notably, our slightly larger MnasNet-A3 model achieves better accuracy than ResNet-50 \cite{resnet16}, but with  \textbf{$\mathbf{4.8\times}$ fewer} parameters and \textbf{$\mathbf{10\times}$ fewer} multiply-add  cost.

\begin{table}
  \centering                                                                        
  \scalebox{0.85}{                                                                  
      \begin{tabular}{c|ccc}                                                     
          \toprule[0.2em]                                                   
                              &   &   Inference Latency & Top-1 Acc. \\
          \toprule[0.1em]
                    \multirow{4}{*}{w/o SE} &  MobileNetV2 & 75ms & 72.0\% \\
                    \vspace{3pt}  &  NASNet & 183ms & 74.0\% \\
                     
                     \cline{2-4}  &  MnasNet-B1 &  77ms & 74.5\% \\  
          \toprule[0.1em]
                    \multirow{2}{*}{w/ SE}    &  MnasNet-A1 & 78ms & 75.2\% \\
					    &  MnasNet-A2 &  84ms & 75.6\% \\  
          \toprule[0.2em]
      \end{tabular}                                                                 
  }                                                                                 
  \caption{                                                                         
      \textbf{Performance Study for Squeeze-and-Excitation SE~\cite{senet18}} -- \TT{MnasNet-A} denote the default MnasNet with SE in search space; \TT{MnasNet-B}  denote  MnasNet with no SE in search space. 
  }                                                                              
  \label{tab:nose}                                                         
\end{table}  
Given that squeeze-and-excitation (SE \cite{senet18}) is relatively new and many existing mobile models don't have this extra optimization, we also show the search results without SE in the search space in Table \ref{tab:nose}; our automated approach still significantly outperforms both MobileNetV2 and NASNet.

\begin{table*}
  \centering                                                                        
  \resizebox{0.98\textwidth}{!}{                                                               
      \begin{tabular}{c|cc|cccc|c}                                                     
          \toprule[0.2em]                                                           
          Network                           &  \#Params   & \#Mult-Adds & $mAP$ & $mAP_S$  & $mAP_M$  & $mAP_L$  & Inference Latency  \\      
          \toprule[0.2em]                                                        
          YOLOv2 \cite{yolo17}        & 50.7M & 17.5B  & 21.6  & 5.0 & 22.4 & 35.5 &  -  \\                    SSD300 \cite{ssd16}         & 36.1M & 35.2B  & 23.2 & 5.3 & 23.2 & 39.6  & -  \\          
          SSD512 \cite{ssd16}         & 36.1M & 99.5B  & 26.8 & 9.0 & 28.9 & 41.9  & -  \\          
          MobileNetV1 + SSDLite \cite{mobilenetv117}    & 5.1M  & 1.3B  & 22.2  &-&-&-   & 270ms\\            
          MobileNetV2 + SSDLite \cite{mobilenetv218}   & 4.3M & 0.8B & 22.1  &-&-&- & 200ms  \\
          \toprule[0.2em]
          \bf MnasNet-A1 + SSDLite      & \bf 4.9M  & \bf 0.8B    & \bf 23.0  &\bf 3.8 &\bf 21.7 & \bf 42.0  & \bf 203ms  \\
          \toprule[0.2em]  
          
      \end{tabular}                                                                 
  }                                                                                 
  \caption{                                                                         
      \textbf{Performance Results on COCO Object Detection} -- \TT{\#Params}: number of trainable parameters; \TT{\#Mult-Adds}: number of multiply-additions per image; $mAP$: standard mean average precision on test-dev2017; $mAP_S, mAP_M, mAP_L$: mean average precision on small, medium, large objects; \TT{Inference Latency}: the inference latency on Pixel 1 Phone.
  }                                                                              
  \label{tab:ssd}                                  
\end{table*}     
 
\subsection{Model Scaling Performance}

Given the myriad application requirements and device heterogeneity present in the real world, developers often scale a model up or down to trade accuracy for latency or model size.
One common scaling technique is to modify the filter size using a depth multiplier  \cite{mobilenetv117}. For example, a depth multiplier of 0.5 halves the number of channels in each layer, thus reducing the latency and model size. Another common scaling technique is to reduce the input image size without changing the network.

Figure \ref{fig:filter-input} compares the model scaling performance of MnasNet and MobileNetV2 by varying  the depth multipliers  and input image sizes. As we change the depth multiplier from 0.35 to 1.4, the inference latency also varies from 20ms to 160ms. As shown in Figure \ref{fig:filter}, our MnasNet model consistently achieves better accuracy than MobileNetV2 for each depth multiplier.  Similarly, our model is also robust to input size changes and consistently outperforms MobileNetV2 (increaseing accuracy by up to \textbf{4.1\%}) across all input image sizes from 96 to 224, as shown in Figure \ref{fig:input}.

\begin{table}
  \centering                                                                        
  \scalebox{0.85}{                                                                  
      \begin{tabular}{ccccc}
      	 \toprule[0.2em]                                                   
      	      &     Params & MAdds & Latency & Top1 Acc. \\
      	\toprule[0.1em]
              MobileNetV2 (0.35x) &1.66M & 59M & 21.4ms & 60.3\% \\
              MnasNet-A1 (0.35x) & 1.7M &63M & 22.8ms & 64.1\% \\
			  MnasNet-search1     & 1.9M & 65M & 22.0ms & 64.9\% \\
			  MnasNet-search2    & 2.0M & 68M & 23.2ms & 66.0\% \\ 
          \toprule[0.2em]
      \end{tabular}                                                                 
  }                                                                                 
  \caption{                                                                         
      \textbf{Model Scaling vs. Model Search} -- \TT{MobileNetV2 (0.35x) and MnasNet-A1 (0.35x)} denote scaling the baseline models with depth multiplier 0.35; \TT{MnasNet-search1/2} denotes models from a new architecture search that targets 22ms latency constraint.
  }                                                                              
  \label{tab:searchvsscale}                                                         
\end{table}  

In addition to model scaling, our approach also allows searching for a new architecture for any latency target. For example, some video applications may require latency as low as 25ms. We can either scale down a baseline model, or search for new models specifically targeted to this latency constraint. Table \ref{tab:searchvsscale} compares these two approaches. For fair comparison,  we use the same 224x224 image sizes for all models. Although our MnasNet already outperforms MobileNetV2 with the same scaling parameters, we can further improve the accuracy with a new architecture search targeting a 22ms latency constraint.

\subsection{COCO Object Detection Performance}

For COCO object detection \cite{coco14}, we pick the MnasNet models in Table \ref{tab:nose} and use them as the feature extractor for SSDLite, a modified resource-efficient version of SSD \cite{mobilenetv218}.  Similar to \cite{mobilenetv218}, we compare our models with other mobile-size SSD or YOLO  models.

Table \ref{tab:ssd} shows the performance of our MnasNet models on COCO. Results for YOLO and SSD are from \cite{yolo17}, while results for MobileNets are from \cite{mobilenetv218}. We train our  models on COCO trainval35k and evaluate them on test-dev2017 by submitting the results to COCO server. As shown in the table, our approach significantly improve the accuracy over MobileNet V1 and V2. Compare to the standard SSD300 detector \cite{ssd16}, our MnasNet model achieves comparable mAP quality (23.0 vs 23.2) as SSD300  with \textbf{$\mathbf{7.4\times}$ fewer} parameters and \textbf{$\mathbf{42\times}$ fewer} multiply-adds.

\section{Ablation Study and Discussion}

In this section, we study the impact of latency constraint and search space, and discuss MnasNet architecture details and the importance of layer diversity.

\begin{figure}
    \centering
	\begin{subfigure}{0.49\linewidth}                                               
		\centering                                                                  
		\includegraphics[width=\linewidth,keepaspectratio=true]{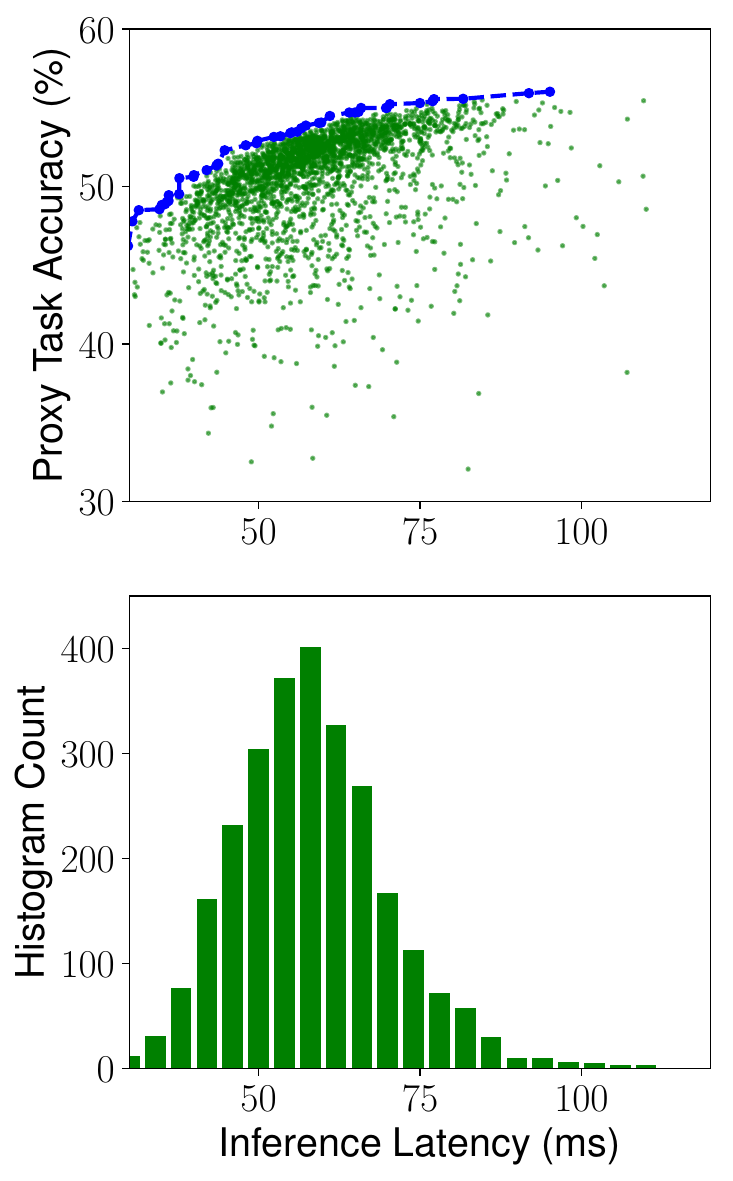}
		\caption{$\alpha=0, \beta=-1$}                  
		\label{fig:hard}
	\end{subfigure}
    \hfill
	\begin{subfigure}{0.49\linewidth}                                               
		\centering                                                                  
		\includegraphics[width=\linewidth,keepaspectratio=true]{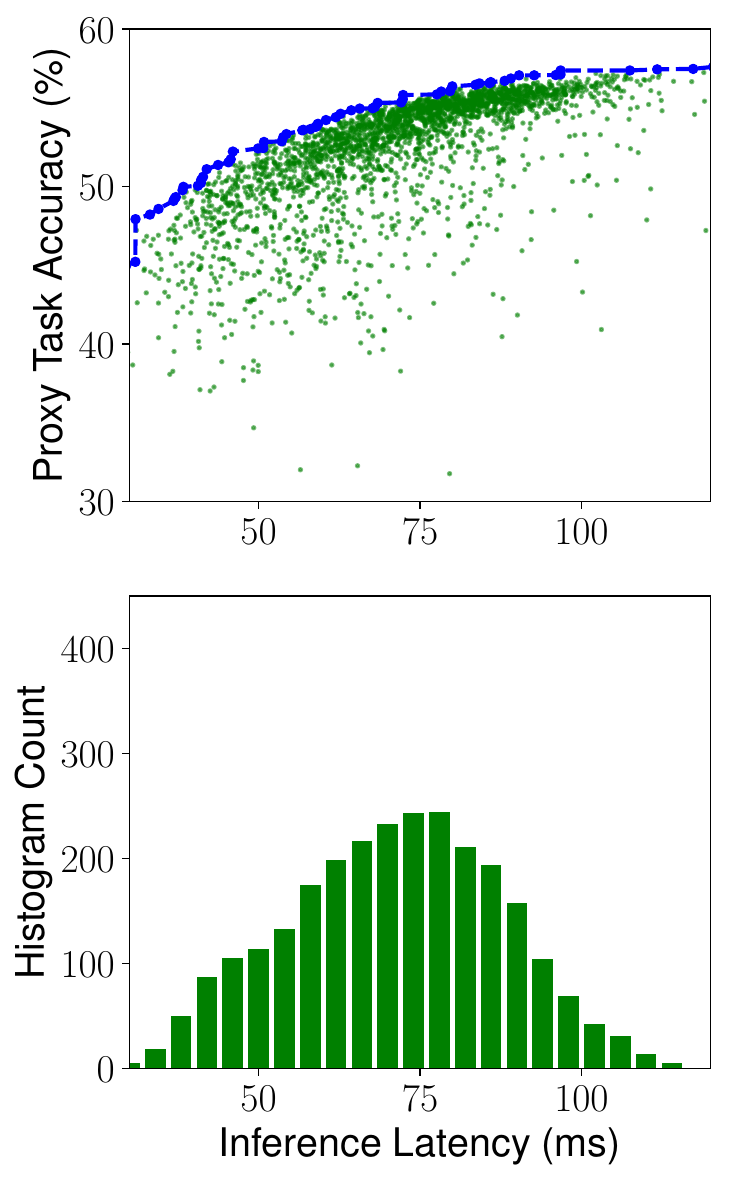}
		\caption{$\alpha=\beta=-0.07$}                  
		\label{fig:soft}
	\end{subfigure}

    \caption{
        \textbf{Multi-Objective Search Results} based on equation \ref{eq:reward} with (a) $\alpha$=0, $\beta$=-1;  and (b) $\alpha$=$\beta$=$-0.07$. Target latency is $T$=$75ms$. Top figure shows the Pareto curve (blue line) for the 3000 sampled models (green dots); bottom figure shows the histogram of model latency.
    }
    \label{fig:pareto}
    \vspace{-0.1in}
\end{figure} \subsection{Soft vs. Hard Latency Constraint}

Our multi-objective search method allows us to deal with both hard and soft latency constraints by setting $\alpha$ and $\beta$ to different values in the reward equation \ref{eq:reward}.
Figure \ref{fig:pareto} shows the multi-objective search results for typical  $\alpha$ and $\beta$. When $\alpha=0, \beta=-1$, the latency is treated as a hard constraint, so the controller tends to focus more on faster models to avoid the latency penalty. On the other hand, by setting $\alpha=\beta=-0.07$, the controller treats the target latency as a soft constraint and tries to search for models across a wider latency range. It samples more models around the target latency value at 75ms, but also explores models with latency smaller than 40ms or greater than 110ms. This allows us to pick multiple models from the Pareto curve in a single architecture search as shown in Table \ref{tab:imagenet}.

\subsection{Disentangling  Search Space and Reward}

To disentangle the impact of our two key contributions: multi-objective reward and new search space, Figure \ref{tab:decouple} compares their performance. Starting from NASNet \cite{nas_imagenet18}, we first employ the same cell-base search space \cite{nas_imagenet18} and simply add the latency constraint using our proposed multiple-object reward. Results show it generates a much faster model by trading the accuracy to latency.  Then, we apply both our multi-objective reward and our new factorized search space, and achieve both higher accuracy and lower latency, suggesting the effectiveness of our search space.
\begin{table}[h]
  \centering
  \resizebox{0.9\linewidth}{!}{
        \begin{tabular}{ll|rr}
        \toprule[0.2em]    
         Reward  & Search Space & Latency & Top-1 Acc.  \\
         \midrule[0.2em]    
         Single-obj \cite{nas_imagenet18} & Cell-based \cite{nas_imagenet18}   &  183ms & 74.0\% \\
         \bf Multi-obj  & Cell-based  \cite{nas_imagenet18}   &  100ms & 72.0\% \\
         \bf Multi-obj  & \bf MnasNet  &  \bf 78ms & \bf 75.2\% \\
         \bottomrule[0.2em]    
         \end{tabular}
  }
  \caption{\BF{Comparison of Decoupled Search Space and Reward Design} -- \BF{Multi-obj} denotes our multi-objective reward; \BF{Single-obj} denotes only optimizing accuracy.}
  \label{tab:decouple}
\end{table} 
\subsection{MnasNet Architecture and Layer Diversity}

\begin{figure}
    \centering
    \includegraphics[width=0.9 \linewidth,keepaspectratio=true]{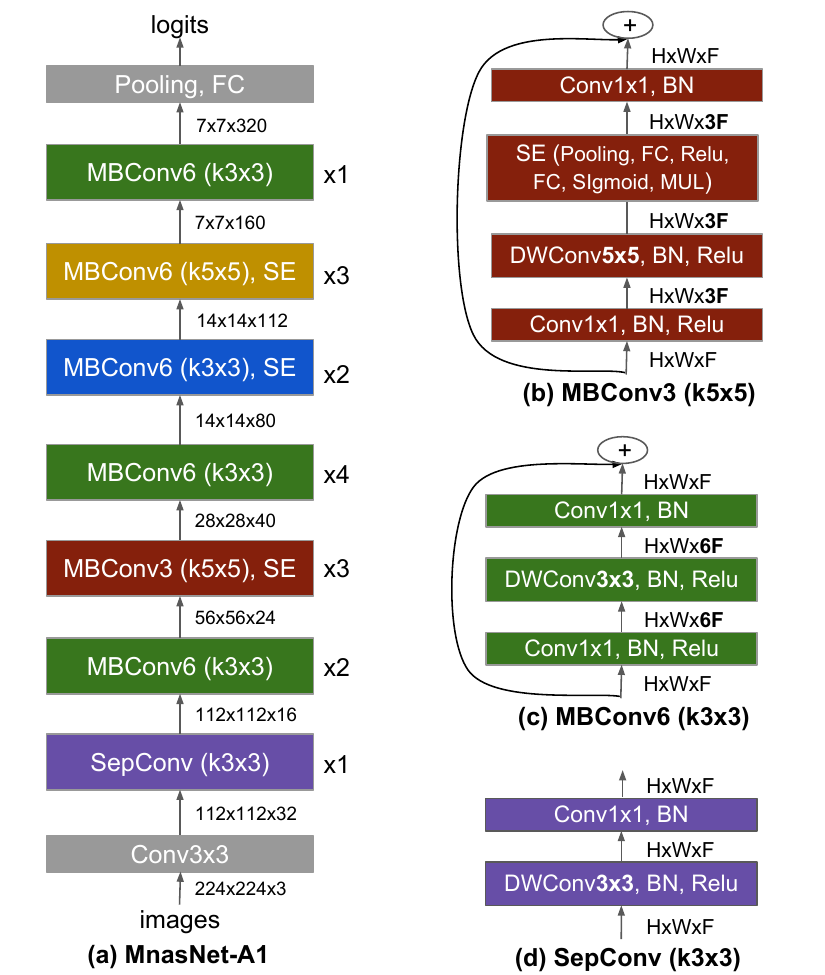}
    \caption{
       \textbf{MnasNet-A1 Architecture} -- (a) is a representative model selected from Table \ref{tab:imagenet}; (b) - (d) are a few corresponding layer structures. \TT{MBConv} denotes mobile inverted bottleneck conv, \TT{DWConv} denotes depthwise conv, k3x3/k5x5 denotes kernel size, \TT{BN} is batch norm, HxWxF denotes tensor shape (height, width, depth), and $\times1/2/3/4$ denotes the number of repeated layers within the block. 
    }
    \label{fig:network}
\end{figure}

Figure \ref{fig:network}(a) illustrates our MnasNet-A1 model found by our automated approach. As expected, it consists of a variety of layer architectures throughout the network. One interesting observation is that our MnasNet uses both 3x3 and 5x5 convolutions, which is different from previous mobile models that all only use 3x3 convolutions.

In order to study the impact of layer diversity, Table \ref{tab:diversity} compares MnasNet  with its variants that only repeat a single type of layer (fixed kernel size and expansion ratio). Our MnasNet model has much better accuracy-latency trade-offs than those variants, highlighting the importance of layer diversity in resource-constrained CNN models. 

\begin{table}[t]
  \centering                                                                        
  \scalebox{0.9}{                                                                  
      \begin{tabular}{c|cc}                                                     
          \toprule[0.2em]                                                           
                                     &  Top-1 Acc. & Inference Latency  \\      
          \toprule[0.2em]                                                        
          \textbf{MnasNet-A1}       & \textbf{75.2\%} & \textbf{78ms}  \\
          \toprule[0.02em]                                                  
          MBConv3 (k3x3) only        & 71.8\% & 63ms  \\
          MBConv3 (k5x5) only        & 72.5\% & 79ms  \\
          MBConv6 (k3x3) only        & 74.9\% & 116ms  \\
          MBConv6 (k5x5) only        & 75.6\% & 146ms  \\          \toprule[0.2em]                                                        
      \end{tabular}                                                                 
  }                                                                                 
  \caption{                                                                         
      \textbf{Performance Comparison of MnasNet and Its Variants} -- \TT{MnasNet-A1} denotes the  model shown in Figure \ref{fig:network}(a); others are variants that repeat a single type of layer throughout the network. All models have the same number of layers and same filter size at each layer.
  }                                                                              
  \label{tab:diversity}                                                         
\end{table}   %
\section{Conclusion}
\label{sec:conclusion}

This paper presents an automated neural architecture search  approach for designing resource-efficient mobile CNN models using reinforcement learning. Our main ideas are incorporating platform-aware \emph{real-world latency information} into the search process and utilizing a novel \emph{factorized hierarchical search space} to search for mobile models with the best trade-offs between accuracy and latency. We demonstrate that our approach can automatically find significantly better mobile models than existing approaches, and achieve new state-of-the-art results on both ImageNet classification and COCO object detection under typical  mobile inference latency constraints. The resulting MnasNet architecture also provides interesting findings on the importance of layer diversity, which will guide us in designing and improving future mobile CNN models.

\section{Acknowledgments}
\label{sec:ack}
We thank Barret Zoph, Dmitry Kalenichenko, Guiheng Zhou, Hongkun Yu, Jeff Dean, Megan Kacholia, Menglong Zhu, Nan Zhang, Shane Almeida, Sheng Li, Vishy Tirumalashetty, Wen Wang, Xiaoqiang Zheng, and the larger device automation platform team, TensorFlow Lite, and Google Brain team.

{\small
	\bibliographystyle{sty/ieee}
	\bibliography{cv}
}

\end{document}